RESEARCH ARTICLE

# Structure-Aware Compound–Protein Affinity Prediction via Graph Neural Networks with Group Lasso Regularization

**Zanyu Shi[1*], Yang Wang[2], Pathum M. Weerawarna[3], Timothy I. Richardson[3], Jie Zhang[4], Yijie Wang[2], and Kun Huang[5*]**

[1]Department of Biostatistics & Health Data Science, Indiana University Fairbanks School of Public Health, Indianapolis, 46202 IN, USA. [2]Department of Computer Science, Luddy School of Informatics, Computing, and Engineering, Indiana University Bloomington, Bloomington, 47405 IN, USA. [3]Division of Clinical Pharmacology, Indiana University School of Medicine, Indianapolis, 46202 IN, USA. [4]Department of Medical and Molecular Genetics, Indiana University School of Medicine, Indianapolis, 46202 IN, USA. [5]Department of Biostatistics & Health Data Science, Indiana University School of Medicine, Indianapolis, 46202 IN, USA.

*Address correspondence to: zanyshi@iu.edu (Z.S.); kunhuang@iu.edu (K.H.)

Explainable artificial intelligence approaches accelerate drug discovery by improving molecular representation learning, identifying key molecular structures, and rationalizing drug property prediction. However, developing end-to-end explainable models for structure–activity relationship modeling in target-specific compound property prediction remains challenging due to the limited availability of compound–protein interaction data for individual targets and the fact that small changes in chemical substituents or local structural motifs can lead to large differences in molecular properties. Thus, optimally leveraging structural and property information and identifying key moieties related to compound–protein affinity for specific targets is essential. We propose a framework implementing graph neural networks (GNNs) to leverage property and structure information from pairs of molecules with activity cliffs targeting specific proteins to predict compound–protein affinity (i.e., half-maximal inhibitory concentration, $IC_{50}$) and explain property differences. To enhance model explainability, we trained GNNs with structure-aware loss functions using group lasso and sparse group lasso regularizations, which prune and highlight molecular subgraphs relevant to activity differences. We applied this framework to the activity cliff data of molecules targeting 6 tyrosine-protein kinases across Src, Abl, and Tec families, as well as anaplastic lymphoma kinase. Integrating common- and uncommon-node information with sparse group lasso improves molecular property prediction for specific protein targets, as evidenced by lower root mean square errors and higher Pearson's correlation coefficients. Applying regularizations also enhances feature attribution for GNNs by boosting graph-level global direction scores and improving atom-level coloring accuracy. These advances strengthen model interpretability in drug discovery pipelines, particularly in identifying critical molecular substructures in lead optimization.

## Introduction

Artificial intelligence accelerates drug discovery with methods such as structure-based virtual screening and quantitative structure–activity relationship (SAR) modeling to learn drug structure information and molecular representation [1]. These methodologies are typically defined as regression or classification models that establish relationships between molecular structure and properties for downstream prediction [2]. Deep learning models such as message-passing neural networks (MPNNs) [3], a subclass of graph neural networks (GNNs) [4], represent molecules as graphs and propagate information along atoms and bonds to learn features that are predictive of molecular properties. Such models support end-to-end prediction tasks, such as drug property prediction, by using the chemical structures of molecular graphs represented in the Simplified Molecular Input Line Entry System (SMILES) format [5]. Beyond 2-dimensional molecular graphs, recent structure-based paradigms increasingly leverage 3-dimensional geometric information and equivariant GNNs to maintain spatial relationships under rotations and translations, particularly in structure generation and diffusion-based modeling [6]. In addition, explainable artificial intelligence (XAI) methods are increasingly adopted in drug discovery to enhance the interpretability of model predictions, facilitate hypothesis generation, and accelerate drug discovery and design by providing transparent insights into model behavior and strengthening user trust in artificial-intelligence-driven systems [7]. Specifically, they can be used to predict drug–protein interactions (DPIs) for cancer [8], determine drug targets in neurological disorders [9], identify potential leads in in silico screening for COVID-19 [10], and predict the potency of compounds for cardiovascular disease [11].

Despite these promising potentials in drug discovery, applying deep learning models to predict SARs faces several challenges. One challenge is data imbalance, which often occurs as chemical databases such as ChEMBL [12] contain a much larger number of inactive compounds than active compounds for a given protein target. Furthermore, numerous subtle alterations in chemical substituents and local structural motifs can markedly impact molecular properties, increase model complexity, and affect model capacity for end-to-end tasks such as drug property prediction. XAI methods have been increasingly applied in such areas to interpret drug property prediction results. For example, Rao et al. [13] presented experimental results demonstrating that XAI methods can provide reliable and informative explanations to chemists in identifying key substructures. However, many previous deep learning approaches that claim their capacity to explain GNN models for graph classification [14–16] often highlight inconsistent molecular substructures, leading to unstable model explainability for drug property prediction.

DPI modeling tasks, such as compound–protein affinity prediction and functional group-based drug property prediction and classification, are important research areas for traditional SAR models and the graph-level tasks of deep learning models like GNNs. These models have been applied to preclinical drug discovery, virtual screening, and lead optimization [2,17]. Recent DPI models span structure-based and structure-free patterns, ranging from 3-dimensional complex-aware models (e.g., MM-DRPNet) [18] to structure-less, pretrained encoder approaches (e.g., DrugForm-DTA) [19], as well as heterogeneous knowledge-graph frameworks that incorporate chemical and proteomic context (e.g., G-PLIP) [20]. Instead of directly inputting imbalanced DPI data into such models, one strategy is to combine molecular activity information such as half-maximal inhibitory concentration ($IC_{50}$) with structure information (e.g., scaffolds and motifs) to integrate datasets. The loss functions for the models are defined based on molecular similarity and potency difference criteria. Activity cliffs (ACs) are generally defined as pairs or groups of structurally similar compounds that are active against the same target but have large differences in potency [21]. A matched molecular pair (MMP) is a pair of compounds that share a common core structure and are distinguished by substituents at a single site [22]. MMP-based AC (MMP-cliff) can be exploited to capture a large difference in potency between the participating compounds by applying the maximum common substructure (MCS) formalism [23] to calculate the molecular scaffold between pairs of compounds binding to a specific target (Fig. 1). For 2 compounds to be considered as sharing a molecular scaffold, the shared substructure must constitute at least a specified fraction of each molecule's framework (i.e., percentage thresholds of shared atoms between the matched pairs). Thus, such AC data can provide critical information regarding key structural components that determine the drastic difference in MMPs. This information can be leveraged to improve drug–protein binding affinity prediction and to provide topology-aware model explanations for structure–activity differences in specific protein targets. In many previous studies [22–24], models were built by training on compound–protein interaction datasets, including various targeting proteins downloaded from cheminformatics databases, to predict compound–protein affinity. While these models achieved reasonable average accuracy over a large number of targets, the accuracy on individual protein targets is often unsatisfactory. By exploiting the information offered by MMP-cliff, we can focus our training on data for a specific protein target.

Besides tasks like drug property prediction, many previous XAI models have explored node- and edge-level tasks in order to identify crucial molecular substructures [13,25] for DPI. As an analytical interpretation of SAR models, color coding aims to assign an importance value for every input feature, such as an atom or a bond in a molecular graph. Such importance values are often visualized through atom or bond coloring, where the structural patterns that drive a prediction are highlighted on the molecular representation of the compound of interest. Feature attribution methods can identify and highlight vital substructures that determine predicted property change by estimating the contribution of each atom in the molecular graphs [26]. When the same substructures are repeatedly highlighted across multiple AC pairs, these patterns can be interpreted by domain experts as chemically meaningful signals, thereby helping users understand how the model associates specific molecular regions with changes in activity [23]. Furthermore, after exploiting various feature attribution techniques (e.g., Gradient × Input [27] and gradient-weighted class activation mapping [Grad-CAM] [28]) to assign importance scores to nodes and edges in the molecular graphs of compound pairs, we can assess feature attribution performance to estimate and compare the accuracy and stability of model explainability under different model architectures, loss functions, and training settings in the context of drug affinity prediction and AC explanation.

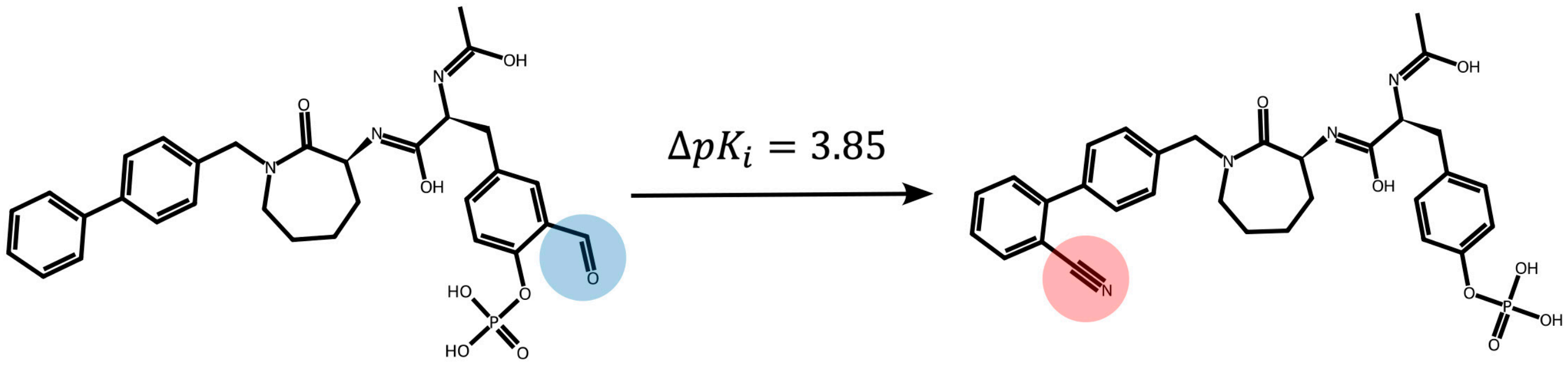

**Fig. 1.** An illustration for an example pair of molecules targeting Src kinase 1042 with activity cliffs ($\Delta pK_i$ = 3.85). The paired 2 molecules share common and uncommon substructures. Common substructures consist of uncolored nodes (atoms) and edges (bonds), and uncommon substituents consist of colored nodes and edges (blue and red circles for sites in the pair).

In this study, unlike approaches that train on large and heterogeneous drug–protein interaction databases, we focus on predicting compound affinity for individual protein families where available DPI data are limited. We applied our framework to predict compound binding affinity (e.g., $IC_{50}$) for tyrosine kinases from the Src [29,30], Abl [31], and Tec (e.g., Bruton's tyrosine kinase [BTK]) [32] families, as well as anaplastic lymphoma kinase (ALK) [33], which are important therapeutic targets in cancer [34] and neurodegenerative diseases such as Alzheimer's disease (AD) [35]. We developed a GNN model framework by leveraging the MMP-cliff information to identify substructures that drive large changes in binding affinity. During training, we incorporated group lasso and sparse group lasso penalties to improve generalization and enhance model performance and explainability.

Overall, our goal is to study how statistical regularization and AC structure information improve both prediction and explainability within a single-target setting with limited compound–protein interaction data. In our framework, each compound is represented as a molecular graph constructed from atoms and bonds, which is further partitioned into 2 complementary subgraphs corresponding to the conserved scaffold and the variable decoration regions. This design reflects the definition of ACs as pairs of structurally similar compounds that share a common core but differ at specific substituent sites that induce large potency changes. By treating scaffold and decoration regions as natural feature groups, group lasso regularization enables the selection or removal of entire subgraphs, thereby encouraging the model to retain informative structural regions while suppressing less relevant ones. This structure-aware regularization links AC learning with chemically meaningful feature selection, supporting both accurate prediction and interpretable structure–activity analysis. Our study leads to the findings below:

- Both common and uncommon substructures derived from MMP-cliff pairs were utilized and integrated to improve model accuracy and enhance model explainability for drug property prediction.
- Loss functions with regularization, including group lasso and sparse group lasso, were used to prune and emphasize informative molecular subgraphs. These regularization methods improve the stability of feature attribution, leading to better atom-level accuracy in node-coloring tasks and higher global direction scores, and thus provide robust subgraph-level explanations.
- Our approach supports model interpretability in drug discovery and virtual screening pipelines, particularly in identifying key molecular substructures that guide lead optimization.

## Methods

In this study, we propose a GNN framework based on the message-passing architecture that leverages structure-aware AC information and incorporates statistical regularization (SAGGLR) to improve binding affinity prediction (Fig. 2). Specifically, we first conducted drug property database screening and generated AC pairs. GNN models were then trained to predict compound–protein affinity (e.g., $IC_{50}$) by learning the molecular representations of both common scaffold regions and uncommon substituent regions in pairs. To improve model performance and identify substructures that drive activity differences, we incorporate regularization strategies, including group lasso and sparse group lasso [36]. Furthermore, we systematically evaluated model performance across different architectures and loss formulations and compared model explainability using both graph-level attribution metrics and atom-level accuracy in node-coloring tasks.

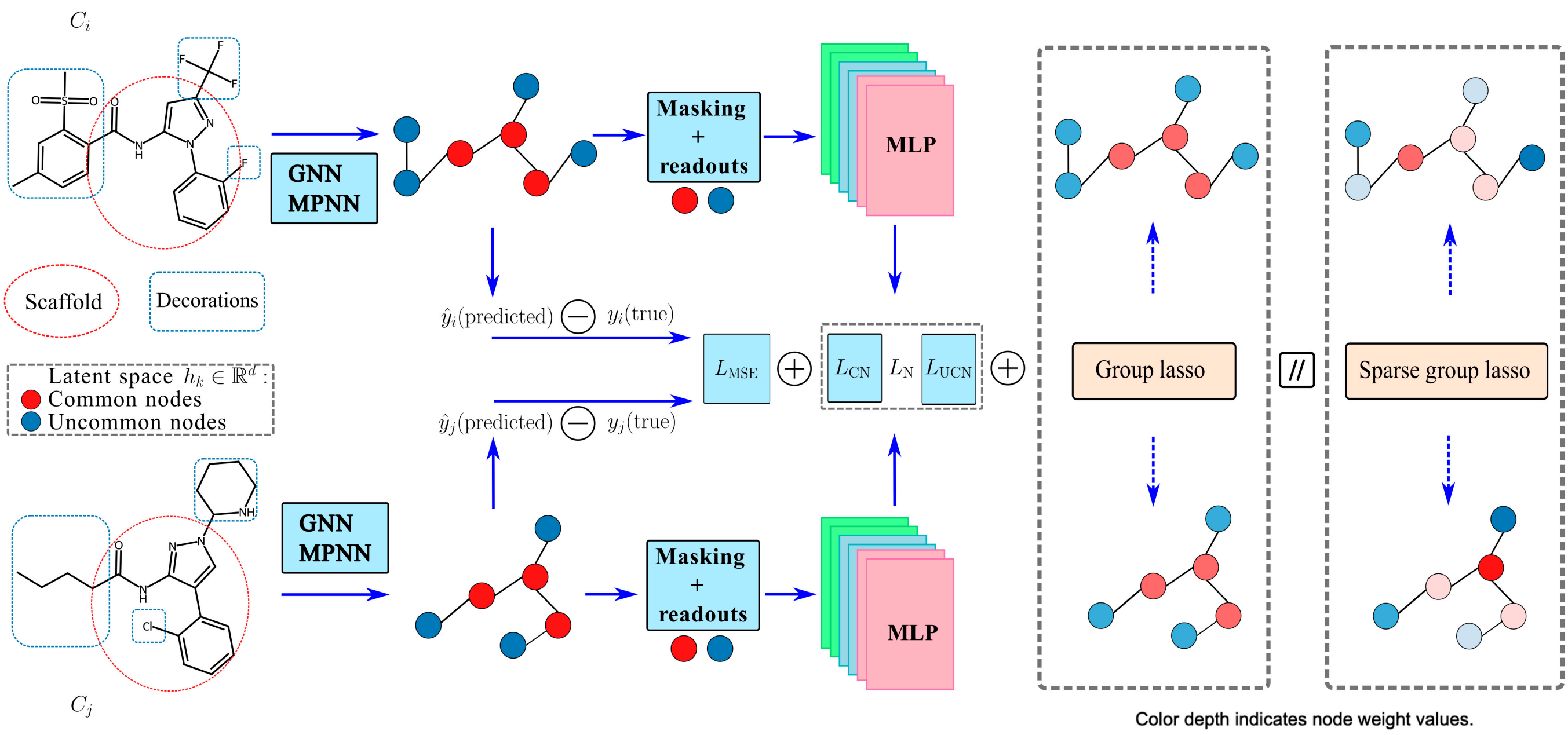

**Fig. 2.** Model structure illustration. Considering a pair of compounds $c_i$ and $c_j$ that share a scaffold in the red circle and decorations in the blue circle, graph neural network (GNN) models and the message-passing neural network (MPNN) framework were applied to learn latent node representations for both common and uncommon nodes. Such node-level information was aggregated to predict graph-level drug–protein binding affinity, and then the mean square errors for predicted and experimental ones were calculated. Additionally, masking functions for common and uncommon nodes and readout functions were applied to normalize the information, and multilayer perceptrons (MLPs) were combined with group lasso and sparse group lasso to create subgraph node loss for both common and uncommon nodes. The 2 loss functions were minimized during the training process. "CN" stands for common nodes and structure; "UCN", for uncommon nodes and structure; and "N", for both common and uncommon nodes. Regularization methods were also applied to prune and select node-level information in activity cliff pairs. Group lasso considers only group-level sparsity (nodes within the same subgroup having the same color depth representing the same weights after pruning), and sparse group lasso considers both group-level and within-group sparsity (nodes within the same subgroup having different color depths representing various weights after pruning).

## Data preparation

For this study, we focused on molecular AC pairs targeting tyrosine kinases, derived from benchmark data reported in a feature attribution study [23], which included 723 protein targets with corresponding small-molecule activity data (i.e., $IC_{50}$). The protein targets and associated small-molecule activities were originally curated from BindingDB validation datasets [37], incorporating $IC_{50}$ values obtained from protein-based assays. Such $IC_{50}$ values stand for drug affinity in this paper's context, and we also defined $pIC_{50} = -\log_{10}(IC_{50})$ for later model training and feature attribution. The MMPs of compounds targeting each protein respectively were curated by the Fast Multiple Maximum Common Substructure (FMCS) algorithm in the RDKit rdFMCS module [23] using the MCS of structurally related compounds exhibiting property cliffs. Specifically, AC pairs can be defined as compound pairs within one or more congeneric series that share a molecular scaffold and exhibit an activity difference in small-molecule activity data (e.g., at least 1 log unit of $pIC_{50}$). We then screened compounds based on such structural similarity criteria to expand the current datasets of tyrosine kinases. In addition, only protein targets with at least 50 compound pairs in the training set were kept. Finally, 3 from the Src family (Protein Data Bank [PDB] IDs [38]: 1O42, 2H8H, and 4MXO), 1 (PDB ID: 2E2B) for ABL1 in the Abl family, 1 (PDB ID: 3AOX) for ALK, and 1 (PDB ID: 3OCS) for BTK in the Tec family were selected from the 25 tyrosine-protein kinase proteins, as well as related compounds and pairs targeting each kinase were provided (Table 1). The Bemis–Murcko scaffold frequencies of compounds targeting each kinase are shown in Tables S1 to S6. The total number of pairs and the number of pairs in training and test sets after splitting for each kinase target in each fold are shown in Table S7.

**Table 1.** Kinase targets and numbers of associated compounds and activity-cliff pairs

| Kinase PDB ID | No. of compounds | No. of activity-cliff pairs |
|---|---|---|
| 1O42 | 90 | 1,433 |
| 2H8H | 63 | 679 |
| 4MXO | 216 | 11,210 |
| 2E2B | 63 | 856 |
| 3AOX | 183 | 5,994 |
| 3OCS | 127 | 2,482 |

PDB, Protein Data Bank

After constructing the data of AC pairs, a 5-fold cross-validation procedure was performed separately on each kinase to evaluate model performance. For each fold, the compound pair dataset was split into a 70% training set, a 10% validation set, and a 20% test set, which were used to train models with different architectures and loss formulations, and to assess model performance and interpretability via feature attribution methods. For each kinase, ligands targeting each kinase were first randomly split into training and test sets within each fold of cross-validation. Although the compound split was first performed at the compound level, the final training and test datasets were constructed at the AC pair level based on the compound assignments. Training and test AC datasets were constructed by retaining only those compound pairs whose first ligand belonged to the corresponding training or test ligand set, thereby generating pair-level splits consistent with the compound-level assignments. Validation sets were later separated from training sets with a given fraction (e.g., 10%).

## Model and loss function design

### *Model framework*

An illustration (Fig. 2) shows the model architecture including GNN layers for molecular representation learning and loss function settings. For any pair $k$ of compounds $i$ and $j$, corresponding to molecular graphs $c_i, c_j \in C$, where $C$ is the graph set of all compound pairs, and drug properties (i.e., $pIC_{50}$) $y_i, y_j \in \mathbb{R}$, we generated and split AC molecular pairs. Next, atom- and bond-level features in molecular graphs were fed into a GNN framework, including atom types such as C (carbon), H (hydrogen), and O (oxygen), and bond features such as encoded bond types (single, double, triple, and aromatic) and stereochemistry (e.g., "STEREOZ", "STEREOE", "STEREOANY", and "STEREONONE") when available. A 50-dimensional node feature vector represented each atom, and each bond was represented by a 10-dimensional edge feature vector. Node and edge features were first projected into a shared 32-dimensional hidden space using separate linear embedding layers.

The GNN backbone consisted of 3 convolutional layers, corresponding to 3 message-passing iterations, each using additive aggregation. In each convolutional layer, the edge network was implemented as a single linear transformation mapping 32-dimensional edge embeddings to a $32 \times 32$ weight matrix (flattened to 1,024 dimensions), following the standard convolutional formulation. Each convolutional layer was followed by batch normalization and ReLU activation. To explicitly model AC effects, common and uncommon atoms between paired molecules were identified and encoded as binary node-level masks. Masked node embeddings were processed by 2 separate linear prediction heads (e.g., $32 \rightarrow 1$) to generate attribution scores for common and uncommon nodes, respectively. In parallel, masked node embeddings were also transformed through 2 linear layers (e.g., $32 \rightarrow 16$) to learn higher-level representations for common and uncommon substructures. These representations were subsequently combined and passed through a final linear layer (e.g., $32 \rightarrow 1$) to produce the predicted binding affinity. Specifically, the GNN layers can be selected from popular message-passing backbones such as NNConv [39,40] with a mean pooling layer, GINConv from graph isomorphism networks (GINs) [41] with an add pooling layer, and GATConv from graph attention networks (GATs) [42] with a mean pooling layer. NNConv is particularly useful for graphs with rich, meaningful edge features (e.g., in computational chemistry where bond types matter) or when the relationship and strength between nodes need to be learned rather than fixed. As regards GINConv, it is excellent for graph classification or node classification tasks where the primary information lies in the node features and structural topology, and edge weights/features are either nonexistent or irrelevant. Furthermore, a modified version of GIN, GINEConv [43], exists to incorporate edge features if needed. In addition, we also used early stopping during model training to mitigate overfitting to avoid possible overestimation of the algorithm's performance. A learning rate scheduler (i.e., torch.optim.lr_scheduler.ReduceLROnPlateau [44]) was adapted using the root mean square error (RMSE) calculated based on the validation sets during the training process. More details about hyperparameter tuning are shown in Appendix A.

### *Loss function design*

For an AC pair $m_k = \left(c_i, c_j\right)$ with true affinities $\left(y_i, y_j\right)$ in a set of $n$ pairs, where $k = 1, \ldots, n$, we defined the true activity difference:

$$\Delta y_k = y_i - y_j \tag{1}$$

Node embeddings $h \in \mathbb{R}^{N \times d}$ were obtained after $L$ message-passing iterations using a chosen GNN backbone (e.g., NNConv, GIN, and GAT) for the graph representation $\left(G_i, G_j\right)$ of compounds in the pair, where $d$ denotes the hidden dimension size in GNN layers:

$$h_i = \mathrm{GNN}\left(G_i\right) \tag{2}$$

$$h_j = \mathrm{GNN}\left(G_j\right) \tag{3}$$

Previous research [23,45] assumed that common substructures are considered neutral, while uncommon structures and their latent space determine the drug property difference. They emphasized the uncommon nodes causing the activity change and proposed a loss scheme focusing on uncommon structural motifs. However, the structural integrity and connections between skeletons and motifs within molecules may collectively influence the drug's properties. To account for this, we applied graph representation learning to jointly learn latent representations of both common and uncommon atoms using tailored loss functions. Specifically, we designed a node-level loss to capture contributions from shared skeletons and uncommon atoms associated with activity changes. To explicitly disentangle shared and differential substructures in AC pairs, common (CN) and uncommon (UCN) atoms were identified via substructure matching and encoded as binary node-level masks. For a molecular graph with $N$ nodes, common and uncommon memberships were represented by binary masks $f^{(\mathrm{CN})}, f^{(\mathrm{UCN})} \in \{0,1\}^N$. A node-level masking operator $F_k(\cdot)$ was applied to produce 2 masks (common and uncommon) by element-wise multiplication:

$$F_k\left(h_i\right) = \left(f^{(\mathrm{CN})} \odot h_i, f^{(\mathrm{UCN})} \odot h_i\right) \rightarrow \left(h_i^{(\mathrm{CN})}, h_i^{(\mathrm{UCN})}\right) \tag{4}$$

$$F_k\left(h_j\right) = \left(f^{(\mathrm{CN})} \odot h_j, f^{(\mathrm{UCN})} \odot h_j\right) \rightarrow \left(h_j^{(\mathrm{CN})}, h_j^{(\mathrm{UCN})}\right) \tag{5}$$

Such masked node embeddings were aggregated into subgraph-level representations subsequently using a pooling layer $R_k(\cdot)$ (e.g., mean, sum, max, or attentional aggregation, depending on the model configuration), producing branch-specific vectors (i.e., common node $g^{(\mathrm{CN})} \in \mathbb{R}^d$ and uncommon node $g^{(\mathrm{UCN})} \in \mathbb{R}^d$):

$$R_k\left(F_k\left(h_i\right)\right) \rightarrow \left(g_i^{(\mathrm{CN})}, g_i^{(\mathrm{UCN})}\right) \tag{6}$$

$$R_k\left(F_k\left(h_j\right)\right) \rightarrow \left(g_j^{(\mathrm{CN})}, g_j^{(\mathrm{UCN})}\right) \tag{7}$$

For each compound, the common-node ($g^{(\mathrm{CN})}$) and uncommon-node ($g^{(\mathrm{UCN})}$) subgraph representations were concatenated into the latent representation of its whole molecular graph. A multilayer perceptron (MLP) layer with a linear output layer was further used to map such latent representation to a scalar affinity prediction, $\hat{y}_k$. The basic loss function (Eq. (8)) was then defined as the mean square error (MSE) to minimize the difference between observed and predicted binding affinities for compounds in pairs during training.

$$L_{\mathrm{MSE}}\left(m_k\right) = \sum_{k \in (i,j)} \left\| y_k - \hat{y}_k \right\|^2 \tag{8}$$

In addition, 2 parallel MLPs were applied to the common-node and uncommon-node subgraph representation, respectively. Each MLP head consisted of a configurable stack of fully connected layers with ReLU activations, followed by a linear projection to a predicted scalar value:

$$\mathrm{MLP}^{(\mathrm{CN})} = \beta^{\mathsf{T}(\mathrm{CN})} g^{(\mathrm{CN})} + b^{(\mathrm{CN})} \tag{9}$$

$$\mathrm{MLP}^{(\mathrm{UCN})} = \beta^{\mathsf{T}(\mathrm{UCN})} g^{(\mathrm{UCN})} + b^{(\mathrm{UCN})} \tag{10}$$

where $\left(\beta^{(\mathrm{CN})}, b^{(\mathrm{CN})}\right)$ and $\left(\beta^{(\mathrm{UCN})}, b^{(\mathrm{UCN})}\right)$ represent the parameters of the common and uncommon prediction heads, respectively. The predicted activity difference can be denoted by the 2 parts:

$$\Delta \hat{y}_k^{(\mathrm{CN})} = \mathrm{MLP}^{(\mathrm{CN})}\left(g_i^{(\mathrm{CN})}\right) - \mathrm{MLP}^{(\mathrm{CN})}\left(g_j^{(\mathrm{CN})}\right) \tag{11}$$

$$\Delta \hat{y}_k^{(\mathrm{UCN})} = \mathrm{MLP}^{(\mathrm{UCN})}\left(g_i^{(\mathrm{UCN})}\right) - \mathrm{MLP}^{(\mathrm{UCN})}\left(g_j^{(\mathrm{UCN})}\right) \tag{12}$$

The node-level loss $L_{\mathrm{N}}$ that describes the contribution of common and uncommon nodes in a molecular pair on the ACs was defined as the sum of the common-structure loss $L_{\mathrm{CN}}$ and the uncommon-structure loss $L_{\mathrm{UCN}}$:

$$\begin{aligned} L_{\mathrm{N}}\left(m_k\right) &= L_{\mathrm{CN}}\left(m_K\right) + L_{\mathrm{UCN}}\left(m_k\right) \\ &= \left\| \lambda^{(\mathrm{CN})} \Delta \hat{y}_k^{(\mathrm{CN})} + \lambda^{(\mathrm{UCN})} \Delta \hat{y}_k^{(\mathrm{UCN})} - \Delta y_k \right\|^2 \end{aligned} \tag{13}$$

where $\lambda^{(\mathrm{CN})}$ and $\lambda^{(\mathrm{UCN})}$ are used to weight the importance of common and uncommon nodes, respectively.

Furthermore, we employed various regularization methods, including group lasso and sparse group lasso penalties, on these loss functions to prune and highlight important molecular substructures. Group lasso is used when features can be naturally grouped into predefined groups, and it selects or discards entire groups of features together rather than individual features [46]. Sparse group lasso extends group lasso by allowing for both group-level sparsity and within-group sparsity [36]. This means that it can further add flexibility to prune and select node-level information for a subset of individual features, given the common and uncommon subgraph groups within AC pairs. For our framework, we added lasso penalty terms in our loss functions as described below:

1. Apply group lasso:

$$L_{\mathrm{MSE}} + L_{\mathrm{N}} + \lambda \left[ \sqrt{p^{(\mathrm{CN})}} \left\| \beta^{(\mathrm{CN})} \right\|_2 + \sqrt{p^{(\mathrm{UCN})}} \left\| \beta^{(\mathrm{UCN})} \right\|_2 \right] \tag{14}$$

where $\beta^{(\mathrm{CN})}$ and $\beta^{(\mathrm{UCN})}$ in Eq. (14) are the parameters for the MLP layers for pruning the common- and uncommon-node information and $p^{(\cdot)}$ is the number of covariates in the common-node subgraph ($p^{(\mathrm{CN})}$) and uncommon-node subgraph ($p^{(\mathrm{UCN})}$).

2. Apply sparse group lasso:

$$L_{\mathrm{MSE}} + L_{\mathrm{N}} + (1-\alpha)\lambda \left[ \sqrt{p^{(\mathrm{CN})}} \left\| \beta^{(\mathrm{CN})} \right\|_2 + \sqrt{p^{(\mathrm{UCN})}} \left\| \beta^{(\mathrm{UCN})} \right\|_2 \right] + \alpha\lambda \|\beta\|_1 \tag{15}$$

where the magnitude of the tuning parameter $\lambda$ determines the sparsity of the solution and $\alpha \in [0, 1]$ denotes a convex combination of the lasso and group lasso penalties ($\alpha = 0$ gives the group lasso fit, while $\alpha = 1$ gives the lasso fit) in Eq. (15).

## Feature attribution

Attribution is an approach to interpretability that highlights the input feature dimensions influential to a neural network's prediction [47]. An attribution map, $G_A = (v_A, e_A)$, is generated by an attribution method $A$ taking a model $M$ and a graph $G$. The attribution map consists of node and edge weights, $v_A, e_A \in \mathbb{R}$, respectively, which are relevant for predicting graph property $y$. These weights can be visualized as a heatmap overlaid on the graph, providing insights into the contributions of different nodes and edges to the prediction.

Feature attribution approaches can help explain which parts of the provided inputs are considered relevant by the underlying supervised learning method for a specific prediction [23]. In the context of drug discovery and molecular design, feature attribution techniques for XAI models typically involve coloring the molecular graphs by assigning a real number of colors to each atom in the graph [26]. Coloring techniques aim to assign an importance value to each input feature, such as an atom or a bond in a molecular graph. These importance values are typically displayed as gradient colorings on atoms or bonds, emphasizing the structural features that drive predictions for the compound's molecular representation. Jiménez-Luna et al. [23] introduced the MCS algorithm to determine proxy ground-truth atom-level color labels for the considered sets. We used this method to determine ground-truth atom-level feature attribution labels via the concept of ACs. Each identified pair constitutes an AC, and we assumed that the observed potency difference for the 2 compounds in a pair results from the structural variations of motifs and their connection with the common scaffolds. Corresponding to the influence of that substructural feature on the predicted property, we labeled molecular subgraphs as negatively influenced and positively influenced nodes with different colors. Features in the more active compound were expected to receive positive attribution, whereas features in the less active compound received negative attribution. Consequently, atomic labels are assigned according to the activity difference sign between the 2 compounds [23]. We obtained the experimental node-level ground truths for attributions, and the edge attributions can be equally redistributed onto their endpoint node attributions.

In this paper, we test multiple feature attribution methods that enable the estimation of positive and negative atom contributions [47], including class activation mapping (CAM) [48], Gradient × Input [27], Grad-CAM [28], and integrated gradients (IG) [49]. For CAM, a global average pooling layer takes node and edge activations and sums them to create a graph embedding layer before outputting to obtain attributions [48]. Gradient × Input computes attributions by taking the input graph's element-wise product and the predicted output's gradient for the input node and edge features. It can also include a reduction step across the feature dimension to produce node- or edge-level attributions [27]. Grad-CAM utilizes the gradient of predicted output for the intermediate activations representing a transformed version of the input to measure input importance and remove the necessity for a global average pooling layer in CAM [28]. IG integrates the element-wise product of an interpolated input with the gradient of the predicted output for the interpolated input, between the actual input and a counterfactual input [49].

## Evaluation and explanation metrics

### *Model performance benchmark*

Considering the comparability of model architecture and dataset formulation, we compared our approaches with 5 representative baselines for molecular property prediction, including a pretraining-based model (i.e., SCAGE [50]) and AC-focused frameworks with specialized architectures (i.e., ACES-GNN [XACs] [51], SemiMol [52], and ACGCN [53]), and a triplet-loss-based method that couples pretraining with AC prediction (i.e., ACtriplet [54]). Model performance was evaluated using RMSE and Pearson's correlation coefficient (PCC) between predicted and true binding affinity ($\mathrm{pIC}_{50}$) values (see Table 2). Cross-kinase paired statistical analyses were conducted by treating each fold–kinase combination as a paired observation ($n = 30$ across 6 kinases, with 5 folds per kinase). Because the distributions of performance differences (RMSE and PCC) did not clearly satisfy normality and the small number of folds per kinase limits reliable normality assessment, we applied both paired $t$ tests [55] and Wilcoxon signed-rank tests [56] as complementary analyses to assess the robustness of our conclusions. In addition, substantial heterogeneity in RMSE levels across kinases led to large between-kinase variability in average RMSE, further motivating the use of both parametric and nonparametric paired tests to ensure that conclusions were not driven by distributional assumptions or outlier kinases.

To test whether thoroughly utilizing molecular graph information and adding loss penalties would improve model performance, we also selected the benchmark model trained with the uncommon-node-focused loss function proposed by Amara et al. [45] and compared it against our models that use loss functions that integrate both common- and uncommon-node information under various regularization schemes (see Table 3). Using the same estimation strategy, the model performance of various model frameworks (i.e., NNConv, GAT, and GIN) with loss penalty settings (i.e., group lasso and sparse group lasso) was also calculated and compared across 6 kinase targets. In addition to cross-kinase fold-level comparisons, we evaluated the consistency of model performance within each kinase. For each of the 6 kinases, fold-level paired comparisons were performed using the 5 cross-validation folds ($n = 5$ per kinase), applying paired $t$ tests and Wilcoxon signed-rank tests to compare model architectures and loss settings. These within-kinase analyses account for repeated cross-validation splits and assess whether performance differences are consistent across folds under each kinase-specific dataset.

**Table 2.** Model performance benchmark evaluation (5-fold cross-validation)

| Model | Avg. RMSE ↓ (±SD) [a] | *P* value [b] |  | Avg. PCC ↑ (±SD) [a] | *P* value [b] |  | Wtd. RMSE ↓ (±SD) [a] | Wtd. PCC ↑ (±SD) [a] |
|---|---|---|---|---|---|---|---|---|
|  |  | Paired *t* test | Wilcoxon |  | Paired *t* test | Wilcoxon |  |  |
| SAGGLR+GL [c] | **0.2665** (±0.1121) | - | - | **0.9545** (±0.0308) | - | - | 0.2115 (±0.0671) | 0.9715 (±0.0187) |
| SAGGLR+SGL [d] | 0.2670 (±0.1180) | - | - | 0.9519 (±0.0402) | - | - | **0.2095** (±0.0671) | **0.9718** (±0.0228) |
| ACGCN | 0.5122 (±0.1595) | $P < 0.001$ | $P < 0.001$ | 0.8761 (±0.0513) | $P < 0.001$ | $P < 0.001$ | 0.4149 (±0.0774) | 0.9215 (±0.0514) |
| ACtriplet | 0.3042 (±0.1371) | $P = 0.117$ | $P = 0.077$ | 0.9515 (±0.0434) | $P = 0.807$ | $P = 0.776$ | 0.2446 (±0.1199) | 0.9711 (±0.0279) |
| ACES-GNN (XACs) | 0.3318 (±0.0916) | $P < 0.001$ | $P < 0.001$ | 0.9526 (±0.0307) | $P = 0.747$ | $P = 0.438$ | 0.3032 (±0.0724) | 0.9605 (±0.0229) |
| SCAGE | 0.7336 (±0.1767) | $P < 0.001$ | $P < 0.001$ | 0.6729 (±0.1647) | $P < 0.001$ | $P < 0.001$ | 0.7079 (±0.1352) | 0.7118 (±0.1135) |
| SemiMol | 0.6205 (±0.1591 | $P < 0.001$ | $P < 0.001$ | 0.8752 (±0.0853) | $P < 0.001$ | $P < 0.001$ | 0.4248 (±0.1280) | 0.9179 (±0.0526) |

[a] Averaged root mean square error (avg. RMSE) and Pearson's correlation coefficient (avg. PCC) across 6 datasets equally and weighted average using their number of activity-cliff pairs in test sets (wtd. RMSE and wtd. PCC) after 5-fold cross-validation. ↓ lower is better; ↑ higher is better. Boldface values show the best model performance.

[b] All paired *t* test and Wilcoxon signed-rank test *P* values were computed by comparing our SAGGLR variant with group lasso (i.e., SAGGLR+GL). "-" stands for working as a baseline (SAGGLR+GL) or excluding from the comparison (SAGGLR+SGL).

[c] SAGGLR+GL represents model setting NNConv + mean with loss setting $L_{\mathrm{N+GL}}$.

[d] SAGGLR+SGL represents model setting NNConv + mean with loss setting $L_{\mathrm{N+SGL}}$.

### *Model interpretability*

Global direction was introduced as a binary metric assessing whether average feature attribution across the uncommon nodes in a pair of compounds preserves the direction of the activity difference [45]. We evaluated the performance of feature attribution methods by using global direction (Eq. (16)). Assuming that $\psi: C \rightarrow \mathbb{R}^{N \times d}$ is a feature attribution function that assigns a score to each node feature in an input graph and the rest of function settings are the same as the loss functions including the readout functions $R(\cdot)$ and the masking function $F(\cdot)$, the computed metric for one pair $m_k = \left(c_i, c_j\right)$ is

$$g_{dir}\left(m_k\right) = \mathbb{1}\left[\mathrm{sign}\left(R_i\left(F_i\left(\psi\left(c_i\right)\right)\right) - R_j\left(F_j\left(\psi\left(c_j\right)\right)\right)\right) = \mathrm{sign}\left(y_i - y_j\right)\right] \quad (16)$$

For each feature attribution method, global direction scores were computed on test-set compound pairs from samples generated under 10 minimum percentages of shared substructure (i.e., MCS thresholds), ranging from 50% to 95% in 5% increments. In addition, we evaluated the stability of atom-level attribution by assessing whether the sign of the attribution assigned to each atom is consistent with the direction of the experimentally observed activity difference between paired compounds. We assigned mask = 0 for common scaffold atoms, mask = 1 for uncommon atoms in the higher-activity compound (i.e., positive ground truth), and mask = −1 for uncommon atoms in the lower-activity compound (i.e., negative ground truth). This enabled a quantitative comparison between predicted attributions and ground-truth labels, and we report standard classification metrics, including accuracy and F1 score, for node-level coloring across different model architectures and loss formulations. In addition, we calculated Spearman's rank correlation [57] between attribution and ground-truth labels to assess whether attribution magnitudes are monotonically aligned with the signed ground-truth directions (±1) on uncommon atoms. A positive Spearman's correlation indicates that atoms whose modifications increase activity tend to receive higher attribution scores than those whose modifications decrease activity. We further evaluated attribution quality using attribution AUROC (area under the receiver operating characteristic curve) [47], with ground-truth direction incorporated by computing AUROC separately for activity-increasing atoms (mask > 0 using raw attributions) and activity-decreasing atoms (mask < 0 using inverted scores, i.e., negative attribution), so that a higher

**Table 3.** Model performance of different models and loss settings (5-fold cross-validation)

| Model (Conv + Pool) | Loss setting ($L_{MSE}$ +) | Avg. RMSE↓ (±SD) [a] | P value [b] Paired t test | P value [b] Wilcoxon | Avg. PCC↑ (±SD) [a] | P value [b] Paired t test | P value [b] Wilcoxon | Wtd. RMSE↓ (±SD) [a] | Wtd. PCC↑ (±SD) [a] |
|---|---|---|---|---|---|---|---|---|---|
| NNConv + mean | $L_{UCN}$ [c] | 0.3284 (±0.1335) | - | - | 0.9256 (±0.0513) | - | - | 0.2646 (±0.0774) | 0.9509 (±0.0296) |
| | $L_{N}$ [d] | 0.3052 (±0.1280) | $P < 0.001$ | $P < 0.001$ | 0.9382 (±0.0458) | $P < 0.001$ | $P < 0.001$ | 0.2423 (±0.0718) | 0.9615 (±0.0256) |
| | $L_{N+GL}$ [e] | **0.2665** (±0.1121) | $P < 0.001$ | $P < 0.001$ | **0.9545** (±0.0308) | $P < 0.001$ | $P < 0.001$ | 0.2115 (±0.0624) | 0.9715 (±0.0187) |
| | $L_{N+SGL}$ [f] | 0.2670 (±0.1180) | $P < 0.001$ | $P < 0.001$ | 0.9519 (±0.0402) | $P < 0.001$ | $P < 0.001$ | **0.2095** (±0.0671) | **0.9718** (±0.0228) |
| GAT + mean | $L_{UCN}$ [c] | 0.3800 (±0.1037) | - | - | 0.9115 (±0.0383) | - | - | 0.3150 (±0.0667) | 0.9374 (±0.0238) |
| | $L_{N}$ [d] | 0.3473 (±0.0874) | $P < 0.001$ | $P < 0.001$ | 0.9327 (±0.0266) | $P < 0.001$ | $P < 0.001$ | 0.2934 (±0.0527) | 0.9517 (±0.0160) |
| | $L_{N+GL}$ [e] | 0.3227 (±0.0949) | $P < 0.001$ | $P < 0.001$ | 0.9365 (±0.0280) | $P < 0.001$ | $P < 0.001$ | 0.2716 (±0.0589) | 0.9544 (±0.0168) |
| | $L_{N+SGL}$ [f] | **0.3055** (±0.0792) | $P < 0.001$ | $P < 0.001$ | **0.9430** (±0.0276) | $P < 0.001$ | $P < 0.001$ | **0.2584** (±0.0513) | **0.9614** (±0.0165) |
| GIN + sum | $L_{UCN}$ [c] | 0.2957 (±0.0996) | - | - | 0.9428 (±0.0356) | - | - | 0.2306 (±0.0669) | 0.9659 (±0.0215) |
| | $L_{N}$ [d] | 0.2778 (±0.0984) | $P < 0.001$ | $P < 0.001$ | 0.9490 (±0.0356) | $P = 0.013$ | $P = 0.015$ | 0.2150 (±0.0633) | 0.9720 (±0.0204) |
| | $L_{N+GL}$ [e] | 0.2437 (±0.0789) | $P < 0.001$ | $P < 0.001$ | 0.9607 (±0.0274) | $P < 0.001$ | $P < 0.001$ | 0.1921 (±0.0519) | 0.9774 (±0.0145) |
| | $L_{N+SGL}$ [f] | **0.2427** (±0.0918) | $P < 0.001$ | $P < 0.001$ | **0.9618** (±0.0269) | $P < 0.001$ | $P < 0.001$ | **0.1878** (±0.0617) | **0.9784** (±0.0160) |

Conv + Pool, convolution and pooling layers; GAT, graph attention network; GIN, graph isomorphism network

[a]Averaged root mean square error (avg. RMSE) and Pearson's correlation coefficient (avg. PCC) across 6 datasets equally and weighted average using their number of activity-cliff pairs in test sets (wtd. RMSE and wtd. PCC) after 5-fold cross-validation. ↓ lower is better; ↑ higher is better. Boldface values show the best model performance across various loss settings under each model.

[b]All paired *t* test and Wilcoxon signed-rank test *P* values were computed by comparing the baseline loss formulation $L_{MSE} + L_{UCN}$ against each alternative loss setting within the same model architecture. "-" stands for working as a baseline.

[c]Loss function only for uncommon substructure (same setting as that in [45]).

[d]Loss function for both common and uncommon substructures.

[e]Loss function with group lasso.

[f]Loss function with sparse group lasso.

AUROC consistently reflects stronger agreement with the ground truth. Perturbation stability was assessed by recomputing Spearman's rank correlation under increasing levels of graph edge dropout rates (i.e., 0%, 5%, 10%, 20%, and 30%), thereby measuring the robustness of attribution quality to structural perturbations.

Because test-set pairs filtered by higher MCS thresholds are nested within those filtered by lower thresholds, observations across thresholds are not entirely independent. Thus, paired statistical tests across thresholds are used primarily to evaluate the consistency and robustness of performance differences, rather than to provide strictly independent hypothesis tests. We therefore applied both paired *t* tests and Wilcoxon signed-rank tests to compare global direction scores, Spearman's correlations, and attribution AUROCs across model architectures and loss functions.

For model explanation, we assumed that the mentioned loss functions designed for GNN models would contribute to the graph-level prediction of the molecular property, and statistical regularization methods could prune and highlight vital molecular substructures at the atom-level molecular property prediction. We first performed explainability evaluation at the graph level by using the global direction scores, which capture both common and uncommon nodes for a compound pair. Thus, we can assess whether the direction of the activity difference is preserved. We then conducted explainability for individual protein targets by mapping feature attribution values to the structures of paired compounds in the context of their binding receptors. An increase in feature attribution performance reflects an improvement in model explainability.

## Results

### Model performance

The model performance benchmark was evaluated via 5-fold cross-validation and compared by using averaged and weight-averaged RMSE and PCC values across the test sets for the 6 kinase targets (reported as mean ± standard deviation in Table 2). Across all benchmarks, our framework variants achieved the best overall accuracy and correlation. Specifically, SAGGLR+GL (i.e., NNConv + mean with $L_{N+GL}$) obtained the lowest average RMSE of 0.2665 and the highest average PCC of 0.9545. In terms of weighted aggregation across targets, SAGGLR+SGL (i.e., NNConv + mean with $L_{N+SGL}$) yielded the lowest weighted RMSE of 0.2095 and the highest weighted PCC of 0.9718. Compared against SAGGLR+GL using paired *t* tests and Wilcoxon signed-rank tests, the majority of baselines (ACGCN, ACES-GNN, SCAGE, and SemiMol) were significantly worse in RMSE and PCC (all $P < 0.001$). ACtriplet was the only baseline without a statistically significant difference from SAGGLR+GL (RMSE: paired *t* test $P = 0.117$ and Wilcoxon signed-rank test $P = 0.077$; PCC: paired *t* test $P = 0.807$ and Wilcoxon signed-rank test $P = 0.776$), despite having a higher mean RMSE (0.3042) and a comparable PCC (0.9510). Overall, these results demonstrate that SAGGLR provides superior predictive accuracy and strong correlation under the single-target benchmark setting, with statistically supported improvements over recent AC-aware baselines.

To test whether more thorough utilization of molecular graph information and the application of loss penalties improve predictive performance, we evaluated GNN models under various combinations of model frameworks (convolution and pooling layers) and loss settings (loss functions and regularization schemes) using 5-fold cross-validation. Model performance was compared by using averaged and weight-averaged RMSE and PCC values across the test sets for the 6 kinase targets (reported as mean ± standard deviation in Table 3) and also their weighted mean values based on the number of pairs in test sets (shown in Table S7). For each model setting (e.g., NNConv + mean, GAT + mean, and GIN + sum), given a basic MSE loss ($L_{MSE}$) defined on the absolute predicted versus experimental binding affinities of molecule pairs, loss function complexity was increased from only adding uncommon-node information ($L_{UCN}$) to considering both common and uncommon substructures ($L_N$) and then applying group lasso ($L_{N+GL}$) or sparse group lasso regularization ($L_{N+SGL}$). The results show an increasingly positive influence on model performance as the loss function complexity increases across all model settings, reflected by the decreasing trend of averaged RMSE values and the increasing trend of averaged PCC values. For NNConv + mean with $L_{N+GL}$, it achieves the lowest averaged RMSE = 0.2665, and the lowest weighted averaged RMSE = 0.2095 and highest weighted averaged PCC = 0.9718 using $L_{N+SGL}$. For GAT + mean with $L_{N+SGL}$, it achieves the lowest averaged RMSE = 0.3055 and lowest weighted averaged RMSE = 0.2584 and the highest averaged PCC = 0.9430 and weighted averaged PCC = 0.9614. For GIN + sum with $L_{N+SGL}$, it achieves the lowest averaged RMSE = 0.2427 and lowest weighted averaged RMSE = 0.1878 and the highest averaged PCC = 0.9618 and weighted averaged PCC = 0.9784.

Paired *t* tests and Wilcoxon signed-rank tests were applied to (a) compare model architectures under fixed loss functions (Table S8 sheet "cross kinase compare models within loss"), (b) compare loss functions under fixed model architectures (Table S8 sheet "cross kinase compare losses within model"), and (c) compare full model-loss configurations (Table 3 and Table S8 sheet "cross kinase compare all loss model pairs"). Wilcoxon signed-rank test *P* values are provided for comparing the baseline loss formulation $L_{MSE} + L_{UCN}$ against each alternative loss setting within the same model architecture (Table 3). These findings demonstrate that drug property prediction is significantly improved by integrating both common- and uncommon-node information and utilizing regularization. For example, compared to that for NNConv + mean with $L_{MSE} + L_{UCN}$, the averaged RMSE for $L_{MSE} + L_{N+GL}$ is reduced by 19.15% (both paired *t* test and Wilcoxon signed-rank test are significant with $P < 0.001$). Compared to that for GAT + mean with $L_{MSE} + L_{UCN}$, the averaged RMSE decreases by 19.60% for $L_{MSE} + L_{N+SGL}$ (both paired *t* test and Wilcoxon signed-rank test are significant with $P < 0.001$). Compared to that for GIN + sum with $L_{MSE} + L_{UCN}$, the averaged RMSE for $L_{MSE} + L_{N+GL}$ is reduced by 17.92% (both paired *t* test and Wilcoxon signed-rank test are significant with $P < 0.001$).

For within-kinase model performance comparison, Wilcoxon signed-rank test *P* values for model performance difference between $L_{MSE} + L_{UCN}$ and other loss settings are shown in Fig. 3, and more test results are reported in Table S8 sheet "performance per kinase fold stat" and sheet "paired tests within kinase". The results indicate consistent trends of decreasing RMSE and increasing PCC as more node information is included, progressing from only uncommon nodes to both common and uncommon nodes. Additionally, increasing the model's complexity by adding more regularization terms, such as transitioning from no penalties to including group lasso and sparse group lasso, leads to significantly improved model performance. This demonstrates that models with regularization outperform those without.

### Model evaluation

The average global direction scores for the loss function $L_{MSE} + L_N$, with and without the group lasso penalty, were compared using scatterplots with connecting lines for each of the 4 feature attribution methods (Fig. 4). Such global direction scores (*y*-axis) were computed and averaged from test-set compound pairs targeting 3 Src family kinases across 10 MCS thresholds (*x*-axis). For graph-level attribution evaluation, we also compared the average global direction scores across 3 Src family kinases using the Wilcoxon signed-rank test, contrasting models trained with the loss $L_{MSE} + L_N$ with and without group lasso regularization. It shows that there is an increase of 47.13% in the global direction score for the feature attribution CAM with a significant Wilcoxon signed-rank test ($P = 0.0002$) (Fig. 4A), an increase of 14.83% in the global direction score for Grad-CAM with a significant Wilcoxon signed-rank test $P = 0.0059$ (Fig. 4B), an increase of 15.4% in the global direction score for Gradient × Input with a significant Wilcoxon signed-rank test ($P = 0.002$) (Fig. 4C), and an increase of 8.49% in the global direction score for IG with a significant Wilcoxon signed-rank test ($P = 0.0098$) (Fig. 4D). The results indicate that across the test datasets and under different MCS thresholds, models trained with a loss function incorporating a group lasso penalty tend to yield higher global direction scores predicted by feature attribution methods than models trained without this penalty. Appendix figures (Figs. B1 to B3) also include scatterplots with connecting lines for compounds targeting 3 Src family kinases respectively, reflecting the increase in global direction scores for $L_{MSE} + L_{N}$ with group lasso compared to the ones without group lasso. The results suggest that introducing regularization like group lasso improves the performance of feature attribution methods.

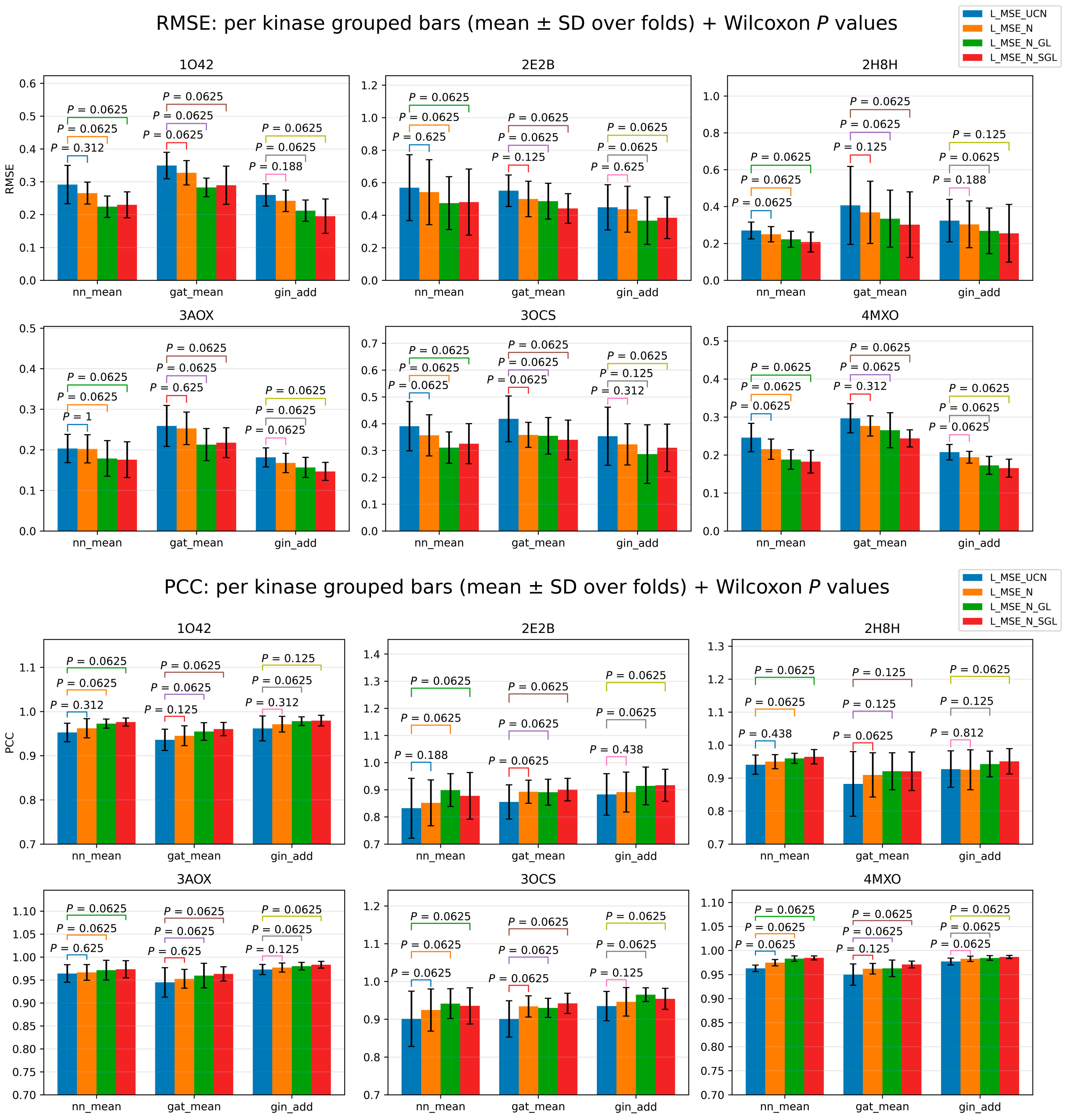

**Fig. 3.** Model performance for the molecules in test sets targeting 6 kinases, respectively, under different model frameworks and loss function settings via 5-fold cross-validation. Each bar plot indicates the averaged root mean square error (RMSE) or Pearson's correlation coefficient (PCC) values for each loss function setting, and error bars represent the standard deviation. Wilcoxon signed-rank test *P* values are provided for $L_{MSE} + L_{UCN}$ compared with other loss settings. The results show a consistent trend of RMSE decreasing and PCC increasing as more node information (from only uncommon nodes to both common and uncommon nodes) and regularization methods were added (from no penalty items for loss functions to adding group lasso and sparse group lasso).

Besides graph-level evaluation, we also conducted quantitative evaluations of node-level coloring predicted by feature attribution approaches for test-set pairs within or across kinases, reporting classification metrics such as accuracy and F1 score across different MCS thresholds, model architectures, and loss formulations in Table S9. For example, given the model architecture NNConv ± mean, the accuracy and PCC values of atom-level coloring prediction for test sets across kinases were significantly improved by implementing group lasso on the loss formulation $L_{MSE} + L_N$ compared to those not, for CAM (with group lasso vs. without penalty and with sparse group lasso vs. without penalty: paired $t$ test and Wilcoxon signed-rank test $P < 0.001$ for both accuracy and F1 scores), Gradient × Input (with sparse group lasso vs. without penalty: paired $t$ test and Wilcoxon signed-rank test $P < 0.001$ for both accuracy and F1 scores), and IG (with group lasso vs. without penalty and with sparse group lasso vs. without penalty: paired $t$ test and Wilcoxon signed-rank test $P < 0.001$ for accuracy). Under the same comparison setting, we further assessed attribution using Spearman's rank correlation between predicted attributions and ground-truth signs, as well as attribution area under the curve [47], and the sensitivity of attributions to graph perturbations. It shows that Spearman's rank correlation was significantly improved for Grad-CAM (with group lasso vs. without penalty and with sparse group lasso vs. without penalty: paired $t$ test and Wilcoxon signed-rank test $P < 0.001$) and IG (with sparse group lasso vs. without penalty: paired $t$ test and Wilcoxon signed-rank test $P < 0.001$ for accuracy). Furthermore, the AUROC for negative nodes (in the low-activity compound of an AC pair) was significantly improved for Gradient × Input (with group lasso vs. without penalty and with sparse group lasso vs. without penalty: paired $t$ test and Wilcoxon signed-rank test $P < 0.001$) and IG (with group lasso vs. without penalty: paired $t$ test $P = 0.017$ and Wilcoxon signed-rank test $P = 0.021$; with sparse group lasso vs. without penalty: paired $t$ test $P = 0.010$ and Wilcoxon signed-rank test $P = 0.090$). In addition, Spearman's rank correlation values after the perturbation stability tests were also significantly improved for CAM (with group lasso vs. without penalty: paired $t$ test and Wilcoxon signed-rank test $P < 0.001$), Grad-CAM (with group lasso vs. without penalty: paired $t$ test and Wilcoxon signed-rank test $P < 0.001$; with sparse group lasso vs. without penalty: paired $t$ test $P < 0.001$ and Wilcoxon signed-rank test $P = 0.055$), and IG (with group lasso vs. without penalty and with sparse group lasso vs. without penalty: paired $t$ test and Wilcoxon signed-rank test $P < 0.001$). More test results are reported in Table S10.

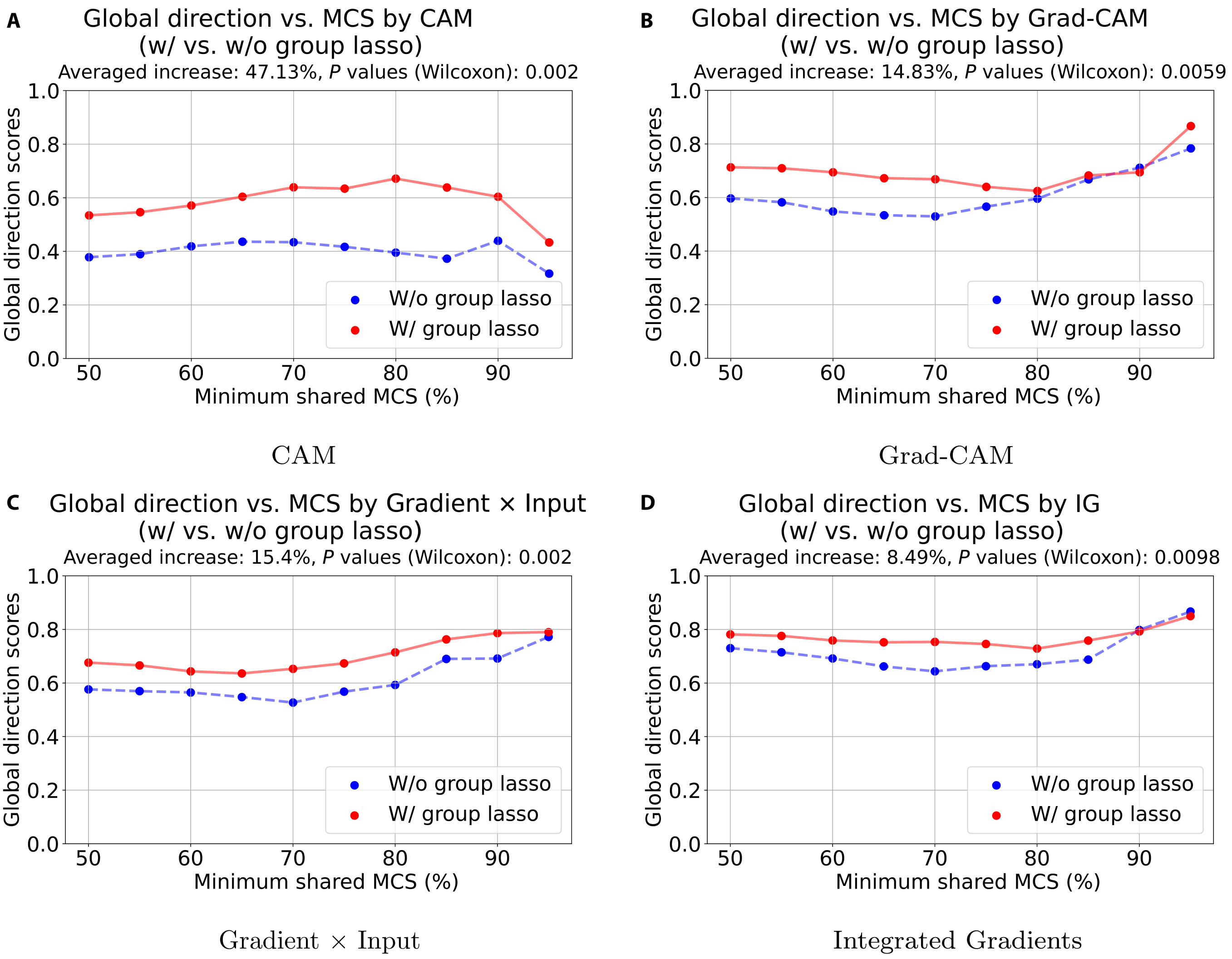

**Fig. 4.** Comparison of averaged global direction scores for $L_{MSE} + L_N$ with and without group lasso by scatterplot with connecting lines. The plot was used for comparing graph-level global direction scores to show the distribution difference of averaged predicted global direction values with $L_{MSE} + L_N$ ($x$-axis) and $L_{MSE} + L_N$ with group lasso ($y$-axis) loss functions under different minimum percentages of shared substructure. Compound pairs are considered at the minimum 50% maximum common substructure (MCS) threshold and from 50% to 100% in 5% increments. The subtitle of each subgraph reports the increased percentage of global direction scores with group lasso regularization for all feature attributions including class activation mapping (CAM) (A) having a 47.13% increase and a significant Wilcoxon signed-rank test ($P = 0.0002$), gradient-weighted class activation mapping (Grad-CAM) (B) having a 14.83% increase and a significant Wilcoxon signed-rank test ($P = 0.0059$), Gradient × Input (C) having a 15.4% increase and a significant Wilcoxon signed-rank test ($P = 0.002$), and integrated gradients (IG) (D) having a 8.49% increase and a significant Wilcoxon signed-rank test ($P = 0.0098$).

### Exemplary explanations for molecules in the test set

A visual comparison was conducted for atom-level accuracy in atom coloring for ground-truth feature attribution labels, atom coloring using the feature attribution approach Grad-CAM with MSE loss and node loss without penalty, and one using Grad-CAM with 2 loss functions with sparse group lasso. Figure 5 shows the structural explanations of one example ligand binding to the 3 Src family kinases 1O42, 2H8H, and 4MXO, respectively, from the test sets. Specifically, the first column shows the ground-truth feature attribution, while the latter 2 columns present the atom coloring captured and predicted by the feature attribution method Grad-CAM with loss function $L_{MSE} + L_N$ with or without penalties. It shows that the predicted coloring of nodes is more consistent for the model trained by using the loss function $L_{MSE} + L_N$ with penalty than that for the model with nonpenalty settings. The results suggest that utilizing regularization, such as sparse group lasso, improves model explainability for graph-level drug property prediction and the identification of vital molecular substructures.

## Discussion

This paper provides a computational and statistical framework for compound–protein affinity prediction by utilizing AC data, exploiting both common- and uncommon-node information to train GNN models, and determining vital substructures by regularization methods such as group lasso and sparse group lasso. It shows that GNN model performance is improved by capturing and integrating both common and uncommon substructure information of AC pairs and relying on loss functions with regularization to prune and highlight molecular subgraphs. It also demonstrates global direction scores and atom-level accuracy in atom coloring predictions are improved by applying regularization methods to enhance feature attribution, thereby providing stable model explainability for molecular property prediction. Moreover, this approach can potentially enhance the interpretability of models in drug discovery and virtual screening pipelines, especially when investigating crucial molecular substructures during lead optimization. Besides drug property prediction and AC explanation, the framework of GNN models combined with statistical regularization methods can also be implemented for other existing GNN approaches that focus on chemical information and intuition, such as key substructure identification for drug synergy prediction [58,59], key substructure learning for material property prediction [60], and drug–drug interaction [61], and molecular relation learning

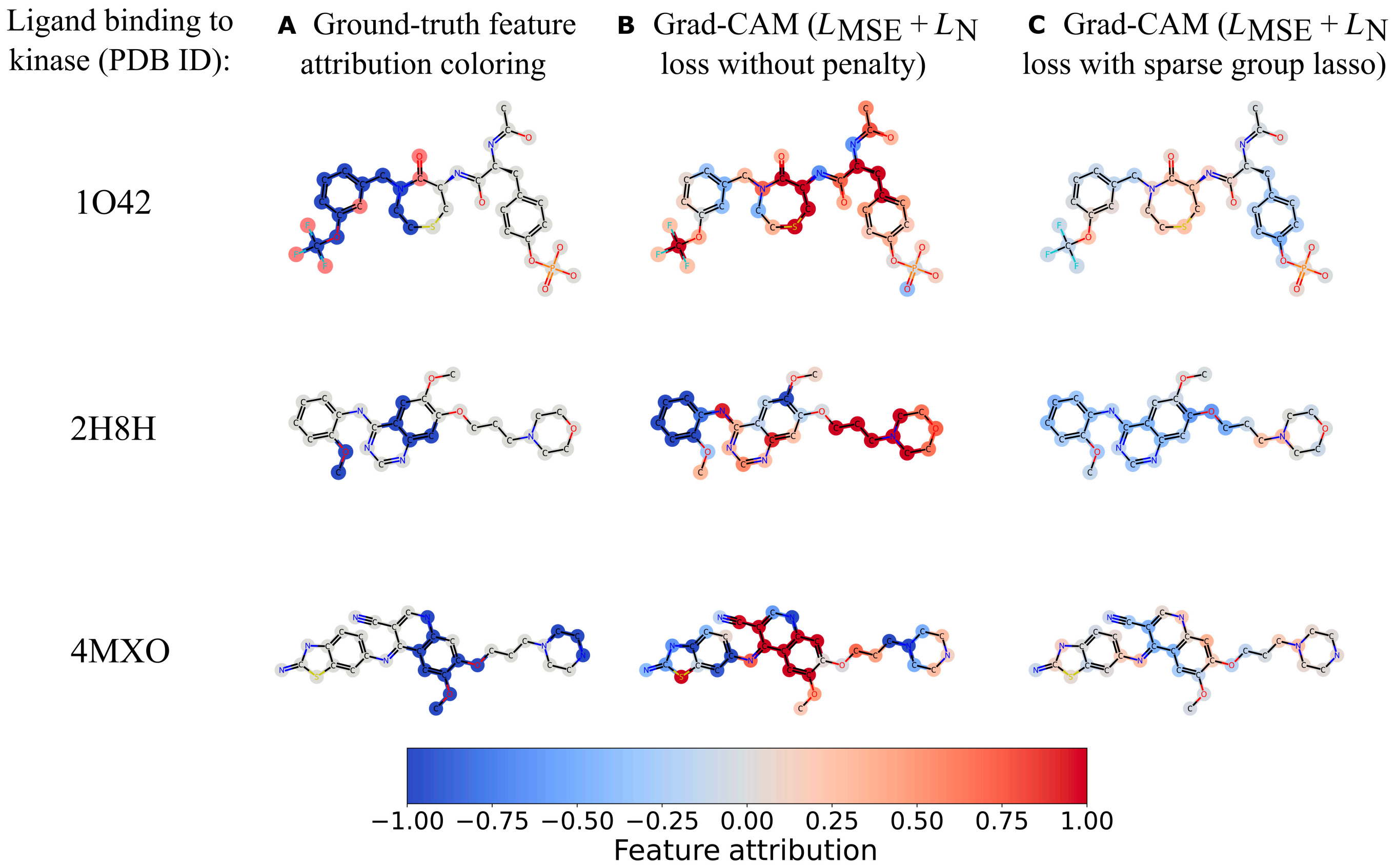

**Fig. 5.** Comparison of the atom-level accuracy in node coloring for ligands binding to the 3 Src family kinases. It compares the (A) ground-truth labels of atom coloring with prediction under 2 other situations: (B) prediction by gradient-weighted class activation mapping (Grad-CAM) for model training with $L_{MSE} + L_N$ without penalty and (C) the one for model training with $L_{MSE} + L_N$ and sparse group lasso. When the sparse group lasso was applied for the loss functions, the predicted atom coloring was much more consistent with the ground-truth feature attribution labels of coloring.

that predicts relationships between molecular pairs by extracting structural features [62,63].

Although this work is primarily methodological, the learned explanations exhibit chemical relevance. The model consistently attributes AC behavior to "uncommon" substructures, which correspond to localized chemical modifications between paired compounds. This aligns with the core principle of SARs, namely, that small, targeted changes at specific functional groups can induce large changes in bioactivity. In contrast, the "common" substructures highlighted by the model tend to form the conserved scaffold shared by compound pairs, which likely represents the core chemotype compatible with tyrosine kinase binding. This pattern suggests that the model is not producing arbitrary explanations but is instead rediscovering chemically intuitive distinctions between scaffold-level compatibility and substituent-level tuning. While we do not claim that the model alone provides mechanistic binding explanations, its ability to localize activity-driving differences offers a practical, data-driven way to guide medicinal chemists toward regions of a molecule that are most promising for optimization in kinase inhibitor design. In addition, limitations of MCS-based attribution exist for its ground-truth assumptions of atom labels [64]. The atom-level attribution labels in this study are derived from MCS differences between MMPs [65], where atoms unique to the more potent compound are assigned positive contributions and those unique to the less potent compound are assigned negative contributions. While this approach is widely adopted for AC and matched-pair analyses and provides a consistent operational reference, it does not constitute definitive causal ground truth. In particular, distal substitutions may influence potency through long-range electronic effects, conformational changes, or scaffold-level interactions [66] that are not captured by local MCS differences, and even modifications within a shared scaffold can alter binding modes or interaction networks. Consequently, discrepancies between model attributions and MCS-based expectations may reflect genuine chemical mechanisms rather than attribution errors. Accordingly, MCS-based labels should be interpreted as a heuristic benchmark for comparative evaluation across large numbers of pairs rather than as atom-level certainty for individual compounds.

Verifying the generality of the model's predictive performance will require extensive computational resources and time for model training with more AC pairs targeting a broader range of proteins. Due to limited time and resources for model training and parameter tuning with different model frameworks and loss settings, we decided to test and compare model performance and evaluate feature attribution approaches on the 6 tyrosine kinase proteins among the Src, Abl, and Tec families, as well as ALK. The Src family has been identified as a promising target for AD therapeutics due to its involvement in key AD-related signaling pathways, including synaptic function and neurodegeneration, and its interaction with amyloid-β and tau pathology, as well as in oncogenic processes such as tumor cell proliferation, survival, migration, and metastasis [35,67]. Abl family kinases, especially the nonreceptor tyrosine kinase c-Abl (ABL1), are implicated in both neurodegeneration and cancer; aberrant c-Abl activation has been observed in AD and models where it promotes neuronal death and tau phosphorylation, and its oncogenic forms drive malignant growth in leukemia and solid tumors [31,68]. Tec family kinases such as BTK and other members play key roles in immune cell signaling and inflammatory responses that contribute to neuroimmune modulation in AD and are established targets in hematologic cancers [32,69]. ALK is an established oncogenic driver whose dysregulation drives tumor growth and survival in several malignancies and has become a therapeutic target in multiple cancers [33]. Emerging evidence also suggests that aberrant activated ALK may promote tau-associated neuronal dysfunction relevant to AD [70]. Collectively, these tyrosine kinases exemplify convergent mechanisms that contribute to both AD pathophysiology and oncogenic signaling, supporting the therapeutic rationale for kinase-targeted strategies across neurodegenerative and cancer contexts.

Utilizing AC pairs instead of compounds targeting individual kinases may mitigate the impact of data imbalance (i.e., inactive compounds outnumber active ones), because there are more available AC pairs than compounds for targeting a specific kinase (Table 1). However, it may also introduce another form of data imbalance, where compounds sharing common scaffolds and motifs with many other compounds appear in numerous AC pairs, potentially affecting model training and prediction performance. Performing random splits on a small, focused dataset is likely to place highly similar compounds in both the training and test sets (Table S7), leading to inflated performance estimates and overly optimistic evaluation results. Although this method employs regularization approaches, such as group lasso and sparse group lasso, to prune and select important common- and uncommon-node substructures, an average pooling layer was used to normalize node information during model training. It may weaken the signals from vital nodes and their substructures, thereby reducing the model's sensitivity in identifying important molecular substructures. This is reflected by the lighter and more even node-coloring prediction compared to the ground-truth feature attribution in Fig. 5. To model how individual substituent sites contribute to AC behavior, we further define loss functions at the node level that link decoration embeddings to experimentally observed activity differences. Another strategy is to integrate explanation-based supervision like Grad-CAM for ACs into GNN training to improve predictive accuracy and interpretability [51], which could extend our current framework.

## Acknowledgments

**Funding:** This work was partially supported by the National Institutes of Health (U54AG065181) and the Indiana University Precision Health Initiative.

**Author contributions:** Z.S.: Conceptualization, data curation, formal analysis, methodology, software, validation, visualization, writing—original draft, and writing—review and editing. Yang Wang: Conceptualization, methodology, and writing—review and editing. P.M.W.: Conceptualization, methodology, and writing—review and editing. T.I.R.: Conceptualization, funding acquisition, project administration, resources, supervision, and writing—review and editing. J.Z.: Conceptualization and writing—review and editing. Yijie Wang: Conceptualization, methodology, supervision, and writing—review and editing. K.H.: Conceptualization, funding acquisition, project administration, methodology, resources, supervision, writing—original draft, and writing—review and editing.

**Competing interests:** The authors declare that they have no competing interests.

## Data Availability

The codes are available at https://github.com/FrancisShizy/SAGGLR.

## Supplementary Materials

Tables S1 to S10

## Appendix A. Hyperparameter Tuning

To ensure fair model comparison and reproducibility, we explicitly describe the hyperparameter selection strategy used for both the GNN backbone and the group lasso regularized objective. All hyperparameters were selected using a nested cross-validation protocol within the training folds, and no test data were used for tuning.

Hyperparameter tuning strategy

For each kinase target, hyperparameters were optimized using 5-fold cross-validation on the training data. A grid-based search was employed for a small, interpretable set of critical hyperparameters rather than exhaustive high-dimensional optimization. Specifically, the regularization strengths for the group lasso and sparse group lasso terms ($\lambda$ and $\alpha$) were selected from logarithmically spaced grids (e.g., $\lambda \in \{10^{-4}, 10^{-3}, 10^{-2}, 10^{-1}\}$ and $\alpha \in \{0.1, 0.3, 0.5, 0.7\}$), with the optimal values chosen to minimize validation RMSE averaged across folds. This strategy balances computational tractability with sufficient coverage of the regularization regime and follows standard practice in penalized regression and graph-based learning. Moreover, the number of neurons in the hidden layers across the model architecture (e.g., embedding, linear, and convolutional layers), as well as the hyperparameter determining the importance of loss function patterns, can also be tuned using the same strategy.

Group lasso specific considerations

The group lasso penalty was applied at the level of predefined node groups corresponding to common and uncommon substructures. The choice of $\lambda$ primarily controls the degree of group-level sparsity (i.e., whether entire substructures are suppressed), while $\alpha$ in the sparse group lasso variant controls within-group sparsity. Empirically, increasing $\lambda$ led to smoother attribution maps and improved generalization for kinases with limited training pairs, whereas overly large $\lambda$ values degraded predictive performance by oversuppressing informative substructures. These observations motivated the use of moderate regularization selected via cross-validation rather than fixed heuristics.

Sensitivity analysis

To assess robustness, we conducted sensitivity analysis with respect to the key hyperparameters, including (a) regularization strengths ($\lambda$ and $\alpha$), (b) message-passing depth (number of GNN layers), and (c) GNN architectural choices (NNConv, GIN, and GAT) and pooling strategies. Performance metrics (RMSE and PCC) were evaluated across a range of values while holding other parameters fixed at their tuned defaults. Across kinases, model performance was stable within a broad neighborhood of the selected hyperparameters, with no abrupt performance collapse observed for small perturbations. In particular, relative comparisons between loss formulations and model variants remained consistent, indicating that finely tuned or brittle parameter choices do not drive the reported conclusions.

Reproducibility and practical applicability

All hyperparameter grids, selected values, and evaluation scripts are provided in the accompanying code repository, enabling exact reproduction of the reported results. While we do not claim global optimality of the chosen hyperparameters, the combination of cross-validated selection, bounded grid search, and sensitivity analysis demonstrates that the proposed framework yields robust performance across reasonable parameter ranges and is practically applicable without extensive manual tuning.

## Appendix B. Global Direction Scores

### Global direction scores for test-set molecules targeting 1042

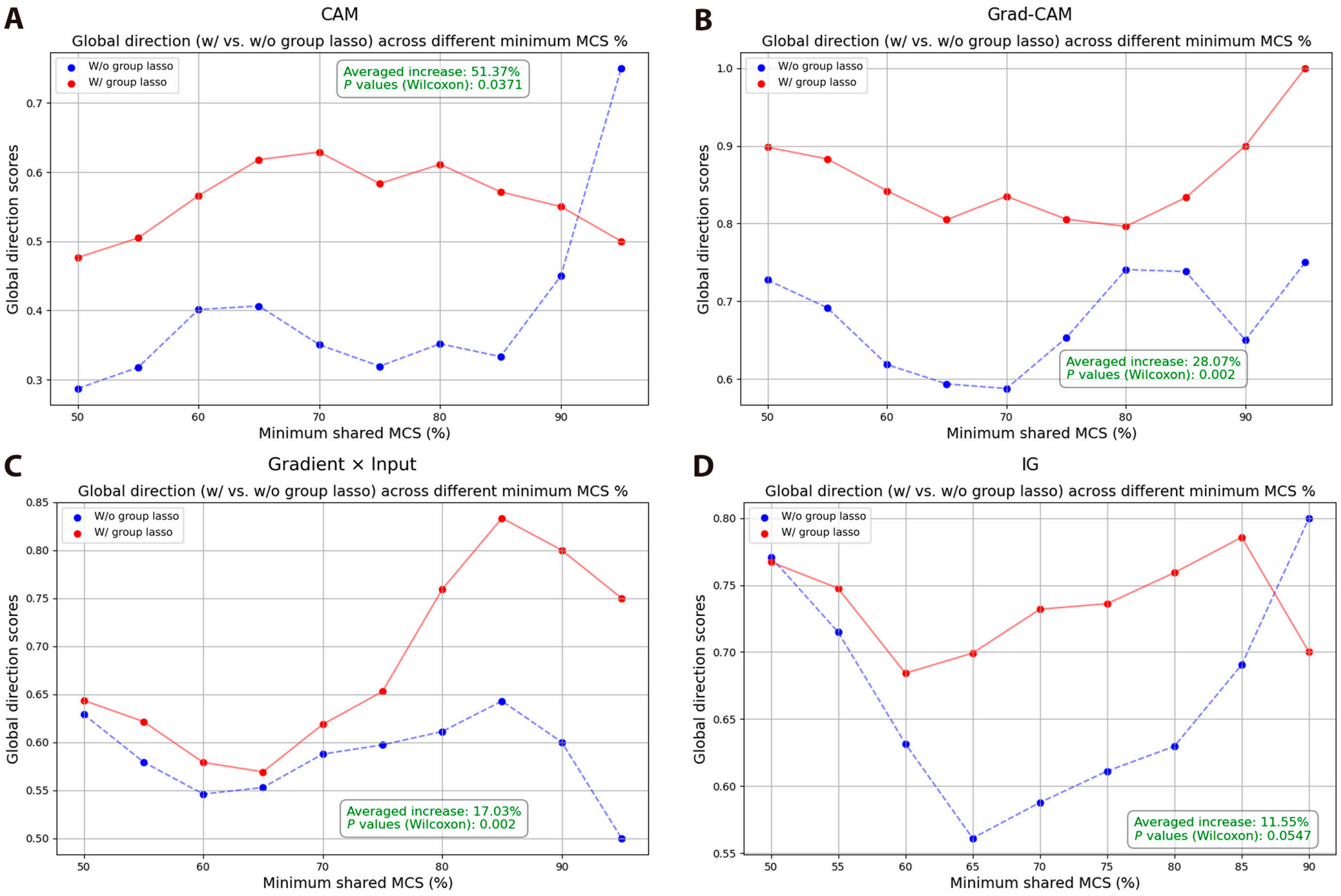

**Fig. B1.** Comparison of global direction scores by using test-set molecules targeting 1042 for $L_{MSE} + L_N$ with and without group lasso by scatterplot with connecting lines. The text box reports the increased percentage of global direction scores with group lasso regularization for all feature attributions including class activation mapping (CAM) (A) having a 51.37% increase and a significant Wilcoxon signed-rank test ($P = 0.0371$), gradient-weighted class activation mapping (Grad-CAM) (B) having a 28.07% increase and a significant Wilcoxon signed-rank test ($P = 0.002$), Gradient × Input (C) having a 17.03% increase and a significant Wilcoxon signed-rank test ($P = 0.002$), and integrated gradients (IG) (D) having a 7.9% increase and a $P$ value of 0.0547.

## Global direction scores for test-set molecules targeting 2H8H

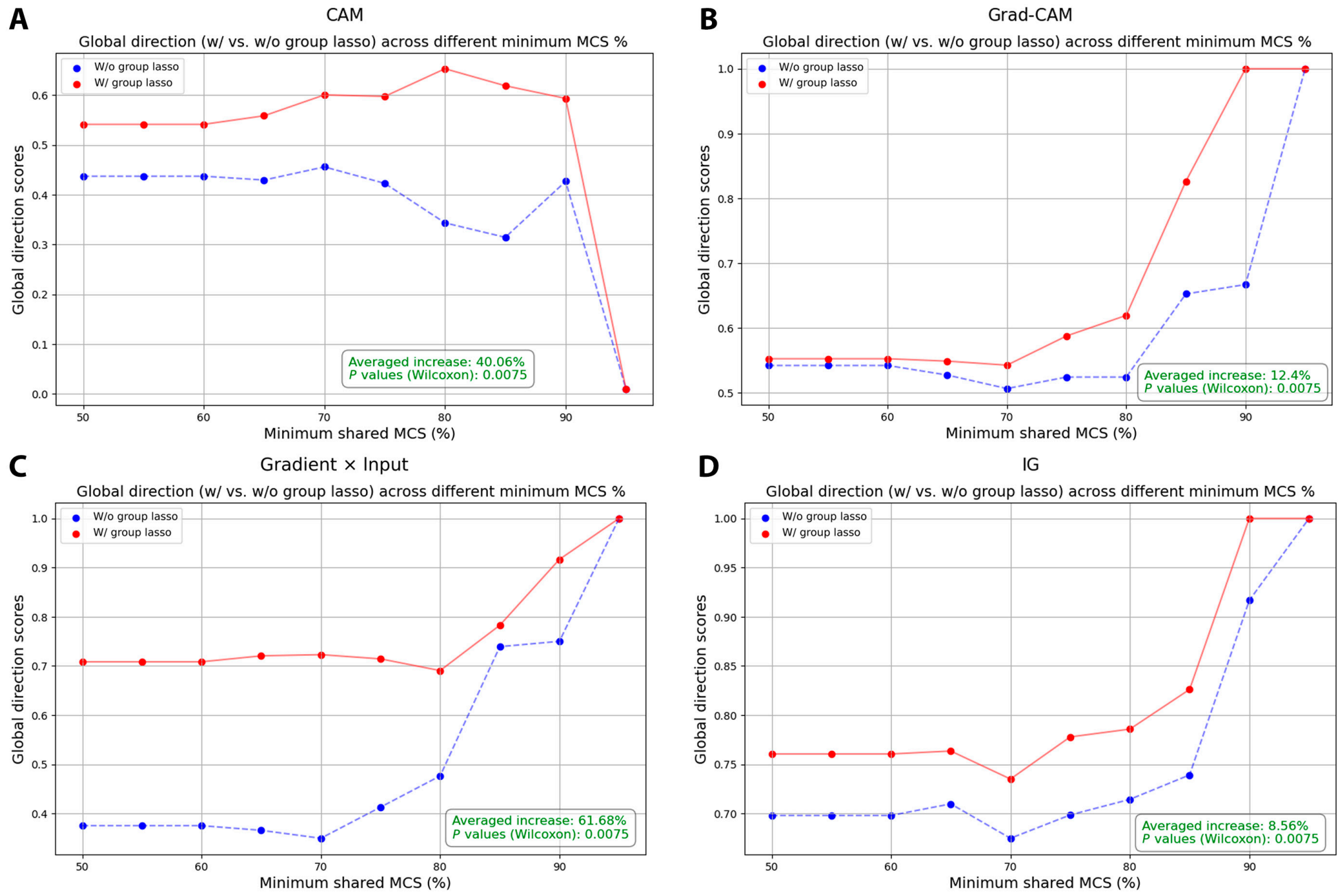

**Fig. B2.** Comparison of global direction scores by using test-set molecules targeting 2H8H for $L_{MSE} + L_N$ with and without group lasso by scatterplot with connecting lines. The text box reports the increased percentage of global direction scores with group lasso regularization for the feature attribution including class activation mapping (CAM) (A) having a 40.06% increase and a significant Wilcoxon signed-rank test ($P = 0.0075$), gradient-weighted class activation mapping (Grad-CAM) (B) having a 12.4% increase and a significant Wilcoxon signed-rank test ($P = 0.0075$), Gradient × Input (C) having a 61.68% increase and a significant Wilcoxon signed-rank test ($P = 0.0075$), and integrated gradients (IG) (D) having a 8.56% increase and a significant Wilcoxon signed-rank test ($P = 0.0075$).

## Global direction scores for test-set molecules targeting 4MXO

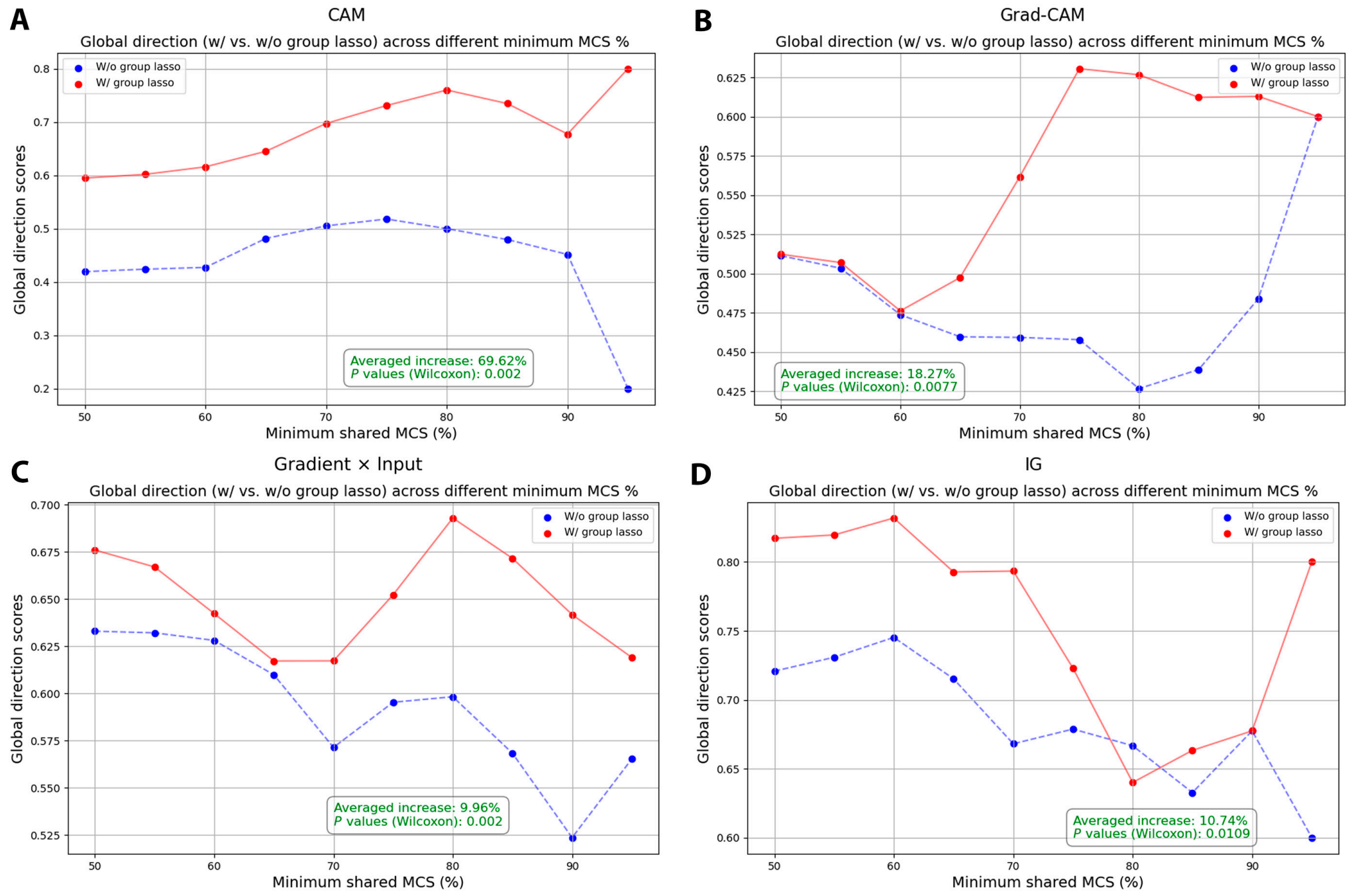

**Fig. B3.** Comparison of global direction scores by using test-set molecules targeting 4MXO for $L_{MSE} + L_N$ with and without group lasso by scatterplot with connecting lines. The text box reports the increased percentage of global direction scores with group lasso regularization for the feature attribution including class activation mapping (CAM) (A) having 69.62% increase and a significant Wilcoxon signed-rank test ($P = 0.002$), gradient-weighted class activation mapping (Grad-CAM) (B) having a 18.27% increase and a significant Wilcoxon signed-rank test ($P = 0.0077$), Gradient × Input (C) having a 9.96% increase and a significant Wilcoxon signed-rank test ($P = 0.002$), and integrated gradients (IG) (D) having a 10.74% increase and a significant Wilcoxon signed-rank test ($P = 0.00109$).

# Structure-Aware Compound–Protein Affinity Prediction via Graph Neural Networks with Group Lasso Regularization

Zanyu Shi, Yang Wang, Pathum M. Weerawarna, Timothy I. Richardson, Jie Zhang, Yijie Wang, and Kun Huang

**Highlights**

- Graph neural networks (GNNs) improve drug property prediction using activity-cliff substructure signals.
- Group lasso regularization enhances atom-level feature attribution in GNN models.
- Sparse regularization provides stable subgraph-level explainability for molecular property prediction.
- The approach enhances the interpretability of GNNs for drug discovery and lead optimization.

Explainable artificial intelligence approaches accelerate drug discovery by improving molecular representation learning, identifying key molecular structures, and rationalizing drug property prediction. However, developing end-to-end explainable models for structure–activity relationship modeling in target-specific compound property prediction remains challenging due to the limited availability of compound–protein interaction data for individual targets and the fact that small changes in chemical substituents or local structural motifs can lead to large differences in molecular properties. Thus, optimally leveraging structural and property information and identifying key moieties related to compound–protein affinity for specific targets is essential. We propose a framework implementing graph neural networks (GNNs) to leverage property and structure information from pairs of molecules with activity cliffs targeting specific proteins to predict compound–protein affinity (i.e., half-maximal inhibitory concentration, $IC_{50}$) and explain property differences. To enhance model explainability, we trained GNNs with structure-aware loss functions using group lasso and sparse group lasso regularizations, which prune and highlight molecular subgraphs relevant to activity differences. We applied this framework to the activity cliff data of molecules targeting 6 tyrosine-protein kinases across Src, Abl, and Tec families, as well as anaplastic lymphoma kinase. Integrating common- and uncommon-node information with sparse group lasso improves molecular property prediction for specific protein targets, as evidenced by lower root mean square errors and higher Pearson’s correlation coefficients. Applying regularizations also enhances feature attribution for GNNs by boosting graph-level global direction scores and improving atom-level coloring accuracy. These advances strengthen model interpretability in drug discovery pipelines, particularly in identifying critical molecular substructures in lead optimization.

**Graphical Abstract**
Graphical abstract
Image

**View the article online**
https://spj.science.org/doi/10.34133/csbj.0012